\documentclass[11pt]{article}

\usepackage[preprint]{acl}

\usepackage{hyperref}
\usepackage{times}
\usepackage{latexsym}
\usepackage[most]{tcolorbox}
\usepackage{listings}
\usepackage{enumitem}

\usepackage[T1]{fontenc}
\usepackage[utf8]{inputenc}
\usepackage{amsmath}
\usepackage{microtype}

\usepackage{inconsolata}

\usepackage{graphicx}
\usepackage{booktabs,makecell,adjustbox,pifont}
\usepackage{fvextra}
\title{Tracing Mathematical Proficiency Through Problem-Solving Processes}

\newcommand*\samethanks[1][\value{footnote}]{\footnotemark[#1]}

\author{
    Jungyang Park\textsuperscript{\rm 1,2}\thanks{~Equal Contribution.}\quad
    Suho Kang\textsuperscript{\rm 3}\samethanks\quad
    Jaewoo Park\textsuperscript{\rm 1}\\
    \textbf{Jaehong Kim\textsuperscript{\rm 2}}\quad
    \textbf{Jaewoo Shin\textsuperscript{\rm 2}}\quad
    \textbf{Seonjoon Park\textsuperscript{\rm 2}}\quad
    \textbf{Youngjae Yu\textsuperscript{\rm 3}}
    \\
    \\
    \textsuperscript{\rm 1}Yonsei University\quad
    \textsuperscript{\rm 2}Mathpresso\quad
    \textsuperscript{\rm 3}Seoul National University \\
    wjddid000624@yonsei.ac.kr \quad youngjaeyu@snu.ac.kr
}

\newcommand{\framework}{\textsc{StatusKT}}
\newcommand{\dataset}{\textsc{KT-PSP-25}}
\newcommand{\task}{\textsc{Knowledge Tracing with Problem-Solving Process}}
\newcommand{\taskabb}{\textsc{KT-PSP}}

\definecolor{customgreen}{HTML}{04BC7D}
\definecolor{customred}{HTML}{EF3333}
\newcommand{\cmark}{\textcolor{customgreen}{\ding{51}}} 
\newcommand{\xmark}{\textcolor{customred}{\ding{55}}}   

\newcommand{\eg}{e.\,g.\ }

\begin{document}

\maketitle

\begin{abstract}
Knowledge Tracing (KT) aims to model student's knowledge state and predict future performance to enable personalized learning in Intelligent Tutoring Systems. However, traditional KT methods face fundamental limitations in explainability, as they rely solely on the response correctness, neglecting the rich information embedded in students' problem-solving processes. To address this gap, we propose Knowledge Tracing Leveraging Problem-Solving Process (KT-PSP), which incorporates students' problem-solving processes to capture the multidimensional aspects of mathematical proficiency. We also introduce KT-PSP-25, a new dataset specifically designed for KT-PSP. Building on this, we present StatusKT, a KT framework that employs a teacher-student-teacher three-stage LLM pipeline to extract students' Mathematical Proficiency (MP) as intermediate representation. In this pipeline, the teacher LLM first extracts problem-specific proficiency indicators, then a student LLM generates responses based on the student's solution process, and a teacher LLM evaluates these responses to determine mastery of each indicator. The experimental results on KT-PSP-25 demonstrate that StatusKT improves the prediction performance of existing KT methods. Moreover, StatusKT provides interpretable explanations for its predictions by explicitly modeling students' mathematical proficiency. Code is available \href{https://github.com/jungyangpark/KT-PSP-25}{here}.
\end{abstract}

\section{Introduction}
\begin{figure}[t]
    \centering
    \includegraphics[width=\linewidth]{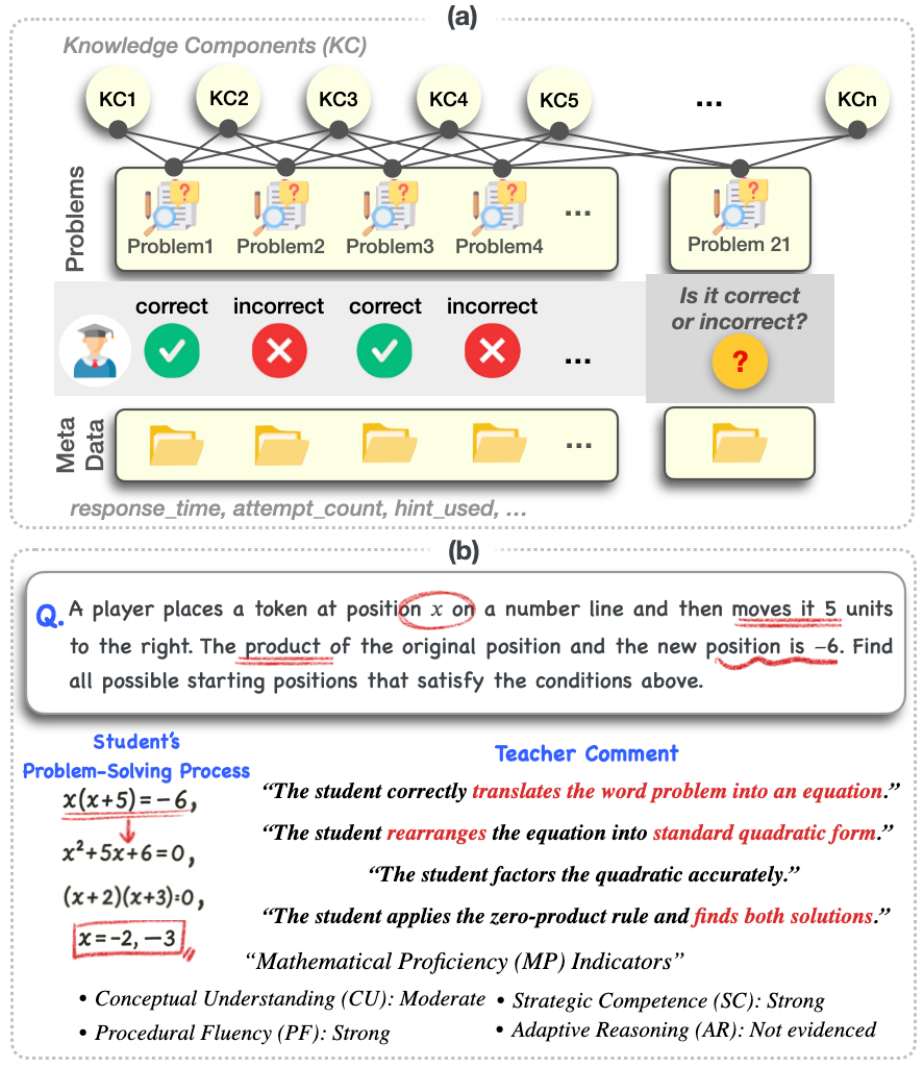}
    \caption{(a) Conventional KT utilizes limited metadata, lacking interpretability. (b) By incorporating problem context, students' reasoning, and teacher comments, \framework\; captures granular proficiency signals aligned with teachers' assessment}
    \label{fig:teaser}\vspace{-1.0em}
\end{figure}

Knowledge Tracing (KT) is a technique that models a learner's evolving knowledge state over time~\citep{corbett1994knowledge, liu2025deep}. Since the true knowledge state cannot be observed directly, KT instead predicts future correctness from historical interaction data, as illustrated in Figure \ref{fig:teaser} (a). Early KT approaches, such as Bayesian Knowledge Tracing (BKT)~\cite{corbett1994knowledge} and Item Response Theory (IRT)~\cite{green1951general}, introduced interpretable probabilistic frameworks but struggled to represent complex learning behavior due to simplifying assumptions (\eg, skill independence and coarse latent dynamics). With the advent of neural networks, deep learning-based KT (DLKT) models have demonstrated notable improvements by employing architectures such as recurrent neural networks~\citep{piech2015deep, nagatani2019augmenting} and attention mechanisms~\citep{pandey2019self, ghosh2020context}. Beyond architectural innovations, subsequent studies have further enriched KT models by integrating diverse contextual signals, such as item difficulty~\citep{yeung2019deep, chen2023improving}, and knowledge concept structure~\cite{su2021time}, thereby enhancing their performance and interpretability.

% Limitations of current task & approaches 
Although KT has advanced in various ways, most approaches rely heavily on outcome-centric supervision (\eg, Correctness) and limited contextual metadata (\eg, Knowledge Concepts (KCs), problem-response time, and difficulty), which often fail to reflect students' underlying understanding~\citep{wang2012leveraging, tschisgale2025exploring, yeung2019deep, liu2024question}. In contrast, as shown in Figure \ref{fig:teaser} (b), real-world teachers evaluate students' understanding through the problem-solving process~\citep{chiu2022analyzing, tschisgale2025exploring, kleinman2022analyzing}. Nevertheless, despite the importance of the solution process, current KT approaches and datasets remain underexplored for incorporating it~\citep{feng2009addressing, liu2023xes3g5m, kim2025kt}.

% Contributions
To address this limitation, we propose \task\ (\taskabb), a new formulation of the KT task that explicitly incorporates students' problem-solving processes (PSP) into the interaction sequence. To facilitate research on \taskabb, we introduce \dataset, a novel dataset that contains real-world PSP for each student-problem interaction. 
%While the original handwritten images cannot be released due to privacy and policy constraints, we provide OCR transcriptions of the solution processes. Alongside these process-level texts, the dataset includes rich problem metadata (KCs, difficulty, question text, and solution text) and interaction attributes (selected answer, duration, and correctness).

Building upon this dataset, we introduce \framework, a process-aware KT framework that uses a three-stage LLM pipeline to extract students' Mathematical Proficiency (MP) from their PSP and use it as an intermediate signal for modeling student knowledge. Inspired by \citet{el2025effect}, \framework\ adopts a teacher-student-teacher structure: a teacher LLM derives problem-specific MP indicators, a student LLM answers them based on the students' written PSP, and a teacher LLM evaluates these answers to produce MP signals. These MP signals are then integrated into the KT backbone as auxiliary inputs to improve predictive accuracy and interpretability. Experiments on \dataset\ demonstrate consistent improvements over strong DLKT baselines and further alleviate the cold-start problem by providing informative MP signals even in early interactions. In addition to a higher predictive accuracy, \framework\ provides interpretable proficiency signals that offer fine-grained insights into students' learning progress. 

The major contributions of this paper are as follows:
\begin{itemize}
    \item We introduce \taskabb, a new KT task formulation that incorporates students' PSP into the interaction sequence, enabling process-aware modeling of knowledge.
    \item We release \dataset, a new mathematical KT dataset containing real-world PSP for each student-problem interaction.
    \item We propose \framework, a novel KT framework that extracts MP signals from students' PSP through teacher-student-teacher LLM pipeline and integrates them as auxiliary representations for KT.
    \item Experiments on \dataset\ show that \framework\ consistently improves prediction performance over DLKT baselines while providing interpretable MP-aligned signals.
\end{itemize}

\section{Related Works}
\subsection{Knowledge Tracing}
\label{Related Works: Knowledge Tracing}
KT plays a crucial role in intelligent tutoring systems by estimating students' evolving knowledge states from their interaction histories. A common line of work infers concept mastery in an autoregressive manner using past question-response data~\citep{piech2015deep, yeung2018addressing, nagatani2019augmenting, guo2021enhancing, zhou2025revisiting}. Memory-augmented models further encode concepts using a static key memory for soft attention and a dynamic value memory to update students' knowledge states~\citep{zhang2017dynamic, abdelrahman2019knowledge}. Graph-based approaches~\citep{nakagawa2019graph, yang2020giktgraphbasedinteractionmodel} capture the structural relations between questions and KCs, propagating historical information via graph updates. Attention-based KT approaches~\citep{pandey2019self, choi2020towards, ghosh2020context, liu2023simplekt, li2024enhancing} improve prediction by selectively focusing on informative past interactions. 

Despite these advances, KT models remain sensitive to noisy behavioral traces. Therefore, recent studies have separated stable cognitive patterns from anomalous outcomes, such as slips~\citep{guo2025enhancing}. However, key determinants remain underexplored, most notably students' PSP and the MP they reveal. Accordingly, we propose a method that leverages fine-grained signals from students' PSP to infer their MP~\citep{findell2001adding, sullivan2011using}, thereby enhancing knowledge tracing.

\subsection{Knowledge Tracing Dataset}
\label{Related Works: Knowledge Tracing Dataset}

The ASSISTments datasets~\citep{feng2009addressing, pardos2014affective} served as early large-scale datasets for KT models. Subsequent datasets, such as Junyi 2015~\cite{chang2015modeling}, extended this line of research by leveraging online learning logs. In parallel, researchers began to evaluate students' performance rather than correctness alone, as reflected in initiatives such as the KDD Cup 2010~\cite{KDD2010}. However, most early datasets contained only simple question-answer sequences, offering little auxiliary information. Over time, KT datasets have expanded beyond mathematics to other domains, such as English, computer science, engineering, and early childhood education~\citep{choi2020ednet, abdelrahman2022dbe, kim2025kt}, motivating richer multimodal and cross-domain learning datasets. Recent efforts by \citet{liu2023xes3g5m} incorporated question content, answer explanations, and hierarchical KC structures to support a more comprehensive evaluation.

However, students' PSP remains largely absent, particularly in mathematical problems, where it is difficult to capture at scale. To bridge this gap, we constructed the first dataset that explicitly recorded PSP.

\section{\task}
\begin{figure*}[t]
    \centering
    \includegraphics[width=\linewidth]{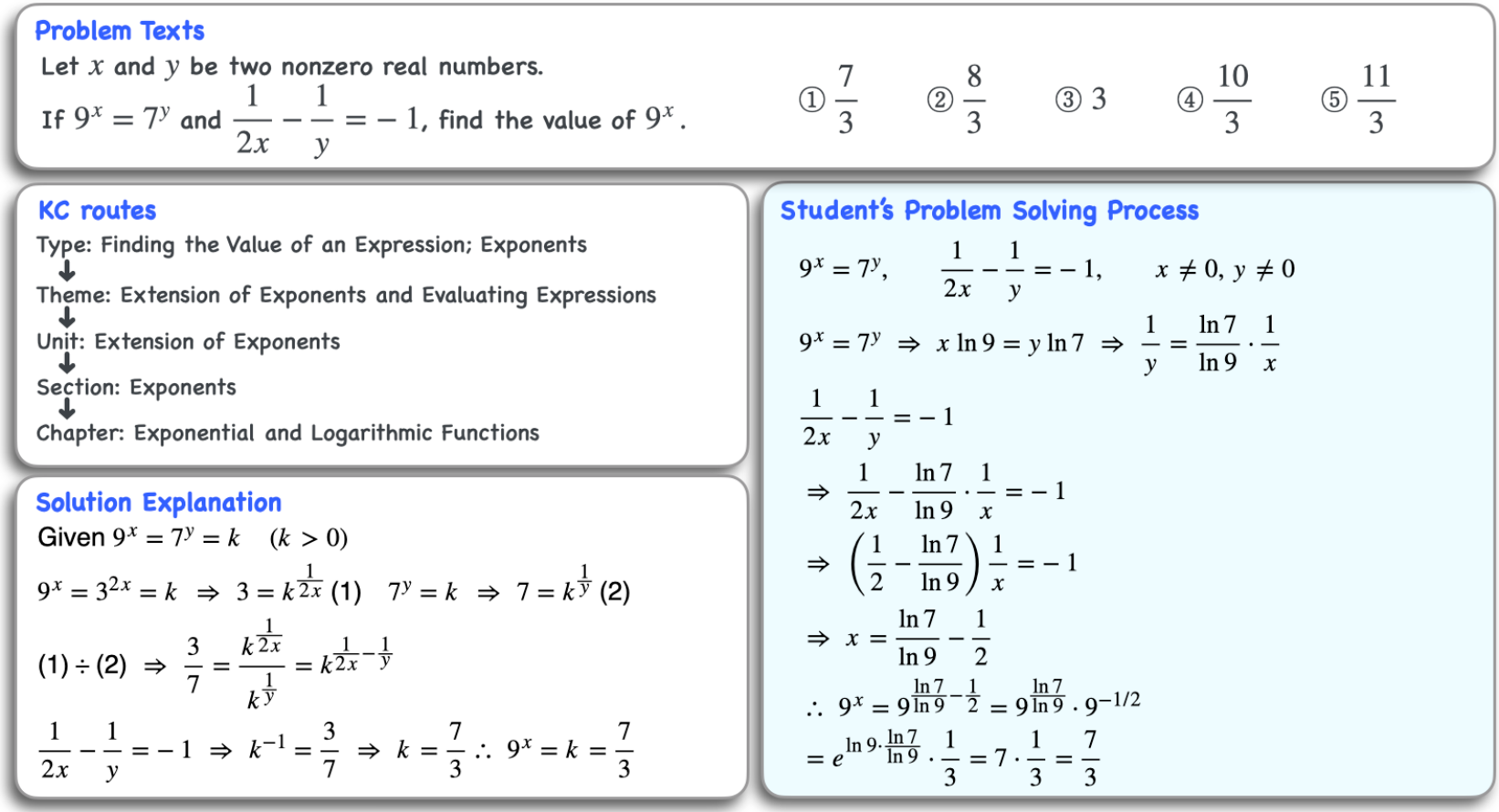}
    \caption{\textbf{An example question from \dataset\ (problem text and KC translated into English).} For short-answer questions, no answer choices are provided.}
    \label{fig:data_example}\vspace{-1.0em}
\end{figure*}

Knowledge Tracing (KT) is a task in educational data mining that aims to model a student's knowledge state over time based on their historical learning interactions. The primary goal is to predict how a student will perform on a future question. Formally, given a student's interaction history sequence 
$\mathcal{S_\textit{conv}}$:

% {\small
% \[
\begin{equation}
\mathcal{S_\textit{conv}} = \{(q_1, c_1, r_1), \dots, (q_t, c_t, r_t)\},
\end{equation}

% \]
% }

where $q_t$ denotes the question at time $t$, $c_t$ is the concept associated with $q_t$, and $r_t \in \{0,1\}$ represents the response correctness (1 for correct, 0 for incorrect), the objective is to estimate $P(r_{t+1} = 1~|~q_{t+1},~ c_{t+1},~ \mathcal{S_\textit{conv}})$.

However, relying solely on response correctness ($r_t$) creates a simplified proxy for students' understanding, often failing to capture the nuances of their actual knowledge state. In real educational environments, students are evaluated not only on the final answer but also on their problem-solving processes (PSP), which provides richer insight into students' understanding~\citep{ukobizaba2021assessment, chiu2022analyzing, tschisgale2025exploring, kleinman2022analyzing}. 

To overcome the limitations of binary outcomes, recent studies have attempted to incorporate auxiliary information into the KT models. 
For instance, response time has been widely used to distinguish between guessing and slipping behaviors~\citep{wang2012leveraging, chen2022introducing, huang2024response}. 
Others have integrated question difficulty or textual content to better estimate the probability of correctness based on item characteristics~\citep{yeung2019deep, liu2024question}. 
% Additionally, interaction features such as hint usage or the number of attempts have been employed to infer a student's struggle level~\citep{rachatasumrit2021toward, xu2023learning}. 
Although these approaches enrich the context, they largely rely on metadata or static attributes, treating the actual PSP as a black box.

% In contrast, in programming education domain, \citet{ross2025modeling} demonstrated that leveraging students' editing traces provides richer signals of reasoning and learning behaviors. However, in the mathematical domain, where the PSP is as critical as the final answer, process-level modeling remains unexplored.

Therefore, we propose \task\ (\taskabb), which extends the traditional KT paradigm by incorporating students' PSP into the interaction history sequence $\mathcal{S_\textit{new}}$:
\begin{equation}
% {\small
% \[
% \mathcal{S_\textit{new}} = \{(q_1, c_1, r_1, p_1), (q_2, c_2, r_2, p_2), \dots, (q_t, c_t, r_t, p_t)\},
\mathcal{S_\textit{new}} = \{(q_1, c_1, r_1, p_1), \dots, (q_t, c_t, r_t, p_t)\},
\end{equation}
% \]
% }

where $p_t$ denotes the detailed PSP for question $q_t$, including step-by-step reasoning, intermediate computations, or other process-level traces. The objective remains estimating $P(r_{t+1} = 1 ~|~ q_{t+1},~c_{t+1},~\mathcal{S_\textit{new}})$, while now leveraging both historical performance and the underlying problem-solving strategies.

\section{Dataset}
\begin{figure*}[t]
    \centering
    \includegraphics[width=\textwidth]{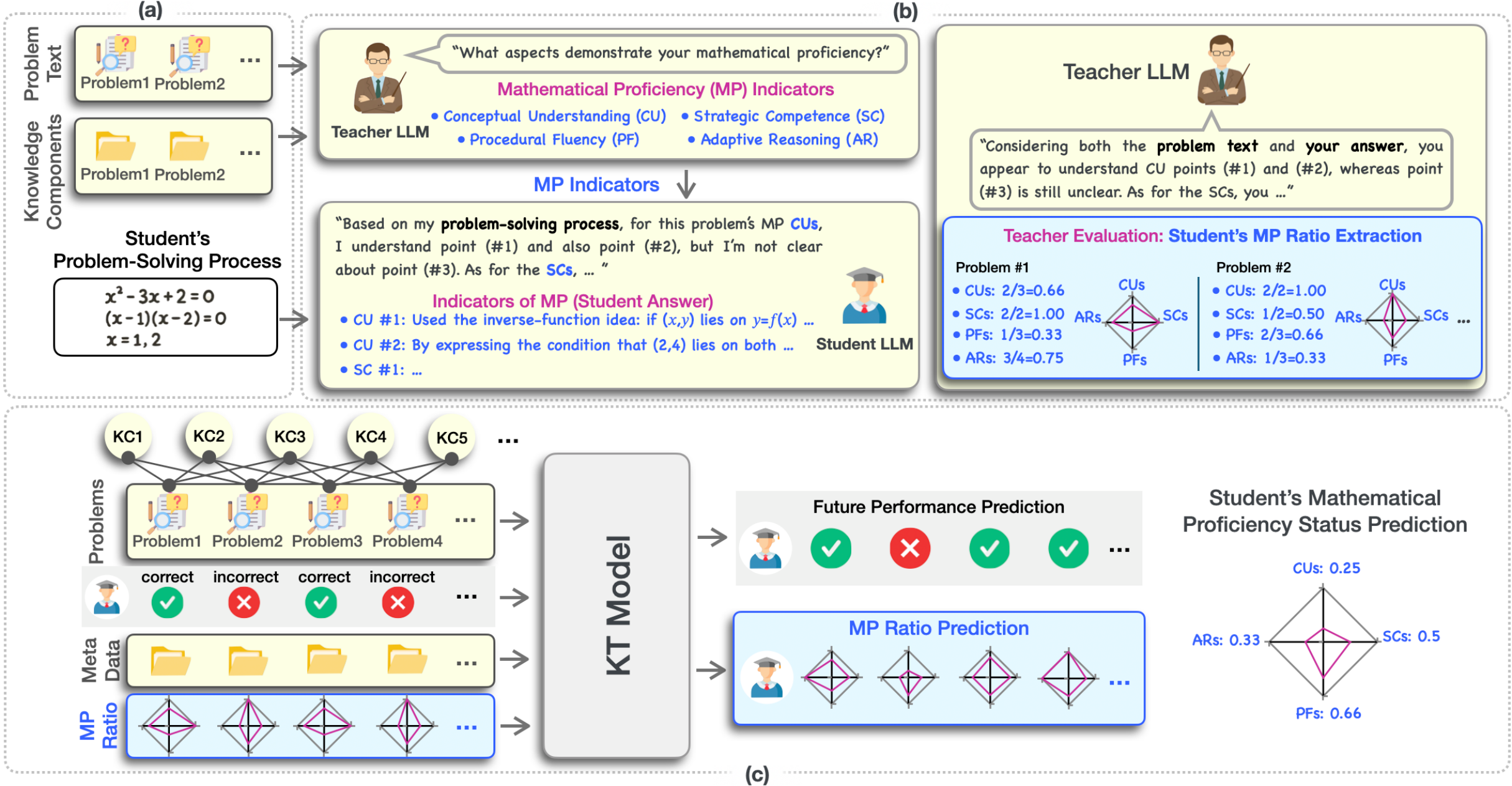}
    \caption{\textbf{\framework\; Framework Overview.} 
    % (a) Students' PSP in math tasks provide rich cues about their understanding, which our framework leverages to derive mathematical proficiency (MP) indicators. 
    (b) A teacher LLM analyzes the text and knowledge components of each problem to construct MP indicators, whereas a student LLM identifies which it understands and can explain. Using these responses, the teacher LLM integrates the results to generate an MP ratio quantifying each student's Mathematical Proficiency(MP). (c) Using the MP ratio, the KT model predicts both students' future performance and problem-level MP ratios, offering a strong rationale for proficiency assessment.}
\label{fig:framework}
\end{figure*}%

\subsection{License}
\dataset\ is released under the CC BY-NC 4.0 (Creative Commons Attribution-NonCommercial 4.0 International) license. This license restricts the dataset to noncommercial use while allowing modifications and sharing under similar terms. Under this license, others can non-commercially remix, adapt, and build upon the work if they credit the source and indicate if changes were made.

\begin{table}[t!]
\centering
\resizebox{\linewidth}{!}{%
\begin{tabular}{l r}
\toprule
Number of interactions & 22,289 \\
Number of students & 1,343 \\
Number of questions & 2,696 \\
Number of Knowledge Components & 490 \\
\midrule

Average solution length (words) & 73.54 \\
Average PSP length (words) & 24.76 \\
Average correct ratio per student & 0.72 \\
Average session duration & 60.36 \\

% \multicolumn{2}{l}{Difficulty distribution} \\
% \quad - Difficulty 10 & 21 \\
% \quad - Difficulty 30 & 507 \\
% \quad - Difficulty 50 & 1,691 \\
% \quad - Difficulty 70 & 444 \\
% \quad - Difficulty 90 & 33 \\

\bottomrule
\end{tabular}
}
\caption{\textbf{Statistics of our dataset}}
\label{tab:dataset_stats}
\end{table}

\subsection{Data Construction}
We curate 22,289 problem-solving sessions from an in-house tablet-based mathematics education platform. Each session corresponds to a single student working on one problem, with interactions recorded between November 2024 and July 2025. To construct the dataset, we applied the following preprocessing steps: (1) remove sessions with fewer than five handwritten lines and (2) discard problems with missing textual content. The resulting dataset, illustrated in Figure~\ref{fig:data_example}, encompasses a wide range of information, including problem-level attributes (problem ID, associated knowledge concepts (KCs), problem text, solution explanation, ground-truth answer, question type, and difficulty) and student interaction attributes (selected answer, duration, PSP, and final correctness).

\subsection{Data Privacy}
Beyond these attributes, \dataset\ also contains rich records of students' interaction sequences, including student IDs, question IDs, and handwritten PSP. Since such data may expose personal information, we implemented privacy-preserving measures. Student and question IDs were mapped to non-reversible digital identifiers to ensure anonymity. 
For handwritten PSP, we applied an OCR pipeline based on GPT-5~\cite{openai2025gpt5}. After initial transcription, we further leveraged GPT to evaluate and refine the OCR outputs, thereby improving both privacy protection and data quality. Detailed prompts and discussion of OCR quality are provided in Appendix~\ref{app:ocr_prompts}
% OCR quality control. We report OCR quality via manual verification on a random subset and analyze how OCR noise affects MP extraction and KT performance in Appendix~\ref{}.
% Reproducibility. We provide prompts, decoding settings, and code for the full preprocessing and MP extraction pipeline.

\subsection{Data Analysis}
There are 22,289 interactions, and 1,343 students answered 2,696 problems from 490 KCs. On average, the reference solution contains 73.54 words while the students' PSP contains 24.76 words, providing substantive reasoning traces rather than short answer-only logs.

Each record additionally stores both the start and completion times, enabling the computation of response durations and temporal analyses across learning sessions. Questions are assigned to five difficulty levels, with most items concentrated around medium difficulty (level 50), while very easy (10) and very hard (90) items appear infrequently. Summary statistics, including the process length, duration, and difficulty breakdown, are reported in Table~\ref{tab:dataset_stats}.

\section{Methodology}
This section presents the details of the proposed \framework, as illustrated in Figure~\ref{fig:framework}. \framework\ enhances knowledge tracing by explicitly modeling students' mathematical proficiency (MP) through interpretable signals derived from their problem-solving process (PSP). While conventional KT models rely on binary correctness, our approach incorporates structured proficiency evidence generated from the PSP. Although \citet{findell2001adding} define MP as comprising five key strands--conceptual understanding (CU), strategic competence (SC), procedural fluency (PF), adaptive reasoning (AR), and productive disposition(PD)--we focus on four observable dimensions, excluding PD because it is difficult to assess reliably through textual problem-solving processes.

\subsection{Deriving MP Indicators from Problem-Solving Processes}
The input to our framework includes OCR-transcribed handwritten solution processes (Figure~\ref{fig:framework} (a)). These traces contain interpretable signals of students' reasoning capabilities. Our pipeline is inspired by the structured assessment protocol in \citet{el2025effect}, in which students respond to a set of predefined MP indicators and mastery is determined based on whether those indicators are satisfied. Similarly, our framework adopts an indicator-based evaluation paradigm. However, because we cannot rely on fixed indicators or direct student interviews, we employ a three-stage Teacher-Student-Teacher LLM pipeline (Figure~\ref{fig:framework} (b)) to automatically extract MP from the raw problem-solving processes:

\textbf{(i). Indicator Extraction(Teacher LLM).} Given the problem $q$ and its associated concept $c$, the first Teacher LLM generates a set of MP indicators:
\begin{equation}
\mathcal{I}_q = f_T(q,c) = \{I_1, \cdots, I_K\},
\end{equation}
where each $I_k$ denotes individual indicators phrased as a question targeting a specific proficiency dimension. These indicators act as a rubric specifying the reasoning elements required for the problem, guiding the subsequent analysis.

\textbf{(ii). Response Generation (Student LLM).} Given the MP indicators $\mathcal{I}_q$ and a student's problem-solving process $p$, the Student LLM generates responses for each indicator:
\begin{equation}
\mathrm{R}_k = f_S(I_k, p).
\end{equation}
This step converts implicit reasoning in the problem-solving process into explicit evidence corresponding to each proficiency dimension.

\textbf{(iii). Proficiency Assessment (Teacher LLM).} Finally, the second teacher LLM evaluates whether each response satisfies its corresponding indicator, producing a binary decision:
\begin{equation}
\mathrm{E}_k = f_T(I_k, R_k), \quad E_k \in \{0,1\}.
\end{equation}

Based on these evaluations, we compute the MP Ratio for each proficiency dimension $d$ as the proportion of satisfied indicators:
\begin{equation}
\mathrm{MP}_d = \frac{1}{|\mathcal{I}_d|} \sum_{k \in \mathcal{I}_d} E_k,
\end{equation}

where $\mathcal{I}_d \subseteq \mathcal{I}_q$ denotes the subset of indicators corresponding to dimension $d$.

In all three stages, we utilized GPT-5~\cite{openai2025gpt5} model. The specific prompts used in the pipeline are provided in Appendix~\ref{app:MP_extraction_prompts}, and the human evaluation protocol and results are summarized in Appendix~\ref{app:MP_Human_Evaluation}.

\subsection{Integrating MP Ratio for Interpretable Knowledge Tracing}
The computed MP ratio serves as an interpretable intermediate representation that complements the traditional KT signals. As depicted in Figure~\ref{fig:framework} (c), the KT backbone utilizes these ratios to predict both (i) students' future performance on subsequent problems and (ii) problem-level MP ratios, enabling fine-grained proficiency tracking. This auxiliary MP prediction encourages the model to internalize proficiency-related patterns, enhancing both the predictive performance and interpretability beyond what correctness alone provides.

To train the model to effectively capture these multidimensional proficiency signals, we defined a composite loss function. The total loss is a weighted sum of the correctness prediction loss and proficiency regression loss:
\begin{equation}
L = \operatorname{BCE}(r_{\text{gt}}, r_{\text{pred}})
   + \alpha \sum_{i \in P} \operatorname{MSE}(m^{\text{gt}}_i, m^{\text{pred}}_i),
\end{equation}
where $P = \{\text{CU}, \text{SC}, \text{PF}, \text{AR}\}$, 
$r$ denotes the response correctness, and 
$m$ represents the MP ratio for dimension $i$. Hyperparameter $\alpha$ controls the trade-off between the correctness prediction and proficiency estimation. 
Through this dual prediction setup, \framework\ improves predictive accuracy while providing interpretable proficiency assessments that align with the established theories of mathematical learning.

\section{Experiments}
\begin{table*}[t]
\centering
\resizebox{\textwidth}{!}{%
\begin{tabular}{lcccccccccc}
\toprule
\textbf{Method} & \textbf{DKT} & \textbf{DKT+} & \textbf{DKVMN} & \textbf{SKVMN} & \textbf{SAKT} & \textbf{SAINT} & \textbf{AKT} & \textbf{simpleKT} & \textbf{stableKT} & \textbf{robustKT} \\
\midrule
\multicolumn{11}{l}{\textit{AUC}} \\
Baseline        & \underline{0.6165} & \underline{0.6192} & \underline{0.6049} & 0.5866 & 0.5819 & 0.6201 & \underline{0.6524} & \underline{0.6591} & \underline{0.6735} & 0.6373 \\
Baseline+\emph{PSP}   & 0.6135 & 0.6169 & 0.6001 & \underline{0.6037} & \textbf{0.6074} & \underline{0.6230} & 0.6457 & 0.6465 & 0.6635 & \textbf{0.6443} \\
\framework      & \textbf{0.6197} & \textbf{0.6200} & \textbf{0.6220} & \textbf{0.6040} & \underline{0.5854} & \textbf{0.6401} & \textbf{0.6629} & \textbf{0.6639} & \textbf{0.6773} & \underline{0.6419} \\
\midrule
\multicolumn{11}{l}{\textit{ACC}} \\
Baseline        & \underline{0.7219} & 0.7226 & 0.7292 & 0.7189 & 0.7243 & 0.7234 & \underline{0.7396} & \underline{0.7393} & \underline{0.7345} & \underline{0.7370} \\
Baseline+\emph{PSP}   & 0.7209 & \underline{0.7233} & \underline{0.7359} & \textbf{0.7335} & \textbf{0.7385} & \underline{0.7326} & 0.7330 & 0.7288 & 0.7219 & 0.7288 \\
\framework      & \textbf{0.7235} & \textbf{0.7247} & \textbf{0.7410} & \underline{0.7257} & \underline{0.7313} & \textbf{0.7377} & \textbf{0.7435} & \textbf{0.7459} & \textbf{0.7421} & \textbf{0.7394} \\
\bottomrule
\end{tabular}
}
\caption{\textbf{Results of the main experiment.} We compare performance across diverse KT architectures under three settings: (i) original baselines, (ii) text-embedded \emph{PSP} variants(which concatenate an encoded problem-solving process with the KT inputs without explicit MP reasoning), and (iii) the proposed \framework\ framework. Bold indicates the best result, and underlining indicates the second best.
% While directly injecting text embeddings yields partial improvements, \framework\ consistently achieves higher AUC and ACC by converting problem-solving processes into structured proficiency signals that more effectively capture students' knowledge states.
}
\label{tab:overall_performance}
\end{table*}

\begin{table*}[t]
\centering
\resizebox{\textwidth}{!}{%
\begin{tabular}{lllllllllll}
\toprule
 & \textbf{DKT} & \textbf{DKT+} & \textbf{DKVMN} & \textbf{SKVMN} & \textbf{SAKT} & \textbf{SAINT} & \textbf{AKT} & \textbf{simpleKT} & \textbf{stableKT} & \textbf{robustKT} \\
\midrule
Paired t-test (p) & 0.0003    & 0.4894   & 2.47e-07  & 0.0004    & 0.0843    & 9.84e-06  & 2.27e-05  & 0.0410    & 0.0018    & 0.0489    \\
Cohen's d         & 1.7888    & 0.2279   & 4.3331    & 1.7144    & 0.6135    & 2.7972    & 2.5221    & 0.7539    & 1.3873    & 0.7196    \\ \hline
\end{tabular}
}
\caption{Statistical significance of AUC improvements (Baseline vs \framework) across 10-fold cross-validation.}
\label{tab:significance_testing}
\end{table*}

\begin{table*}[t]
\centering
\resizebox{\textwidth}{!}{%
\begin{tabular}{lcccccccccc}
\toprule
 & DKT & DKT+ & DKVMN & SKVMN & SAKT & SAINT & AKT & SimpleKT & StableKT & RobustKT \\
\midrule
\multicolumn{11}{l}{\textit{AUC}} \\
Baseline        & \textbf{0.6072} & \textbf{0.6093} & \textbf{0.5875} & 0.5758 & 0.5764 & 0.6228 & 0.6217 & 0.6406 & 0.6516 & 0.6196 \\
\framework      & 0.6031 & 0.6069 & 0.5843 & \textbf{0.5826} & \textbf{0.5818} & \textbf{0.6266} & \textbf{0.6352} & \textbf{0.6442} & \textbf{0.6541} & \textbf{0.6302} \\
\midrule

\multicolumn{11}{l}{\textit{ACC}} \\
Baseline        & 0.7064 & 0.7107 & 0.7043 & 0.7207 & 0.7113 & 0.7350 & 0.7189 & 0.7219 & 0.7308 & 0.7330 \\
\framework      & \textbf{0.7146} & \textbf{0.7134} & \textbf{0.7083} & \textbf{0.7253} & \textbf{0.7230} & \textbf{0.7471} & \textbf{0.7278} & \textbf{0.7261} & \textbf{0.7538} & \textbf{0.7402} \\

\bottomrule
\end{tabular}
}
\caption{\textbf{Cold-start performance (\(t \le 5\)).} We evaluate each KT architecture using only the first five observed interactions per student. Models augmented with \framework\ generally achieve higher AUC and ACC, indicating that MP signals help estimate learners’ knowledge states even under limited early evidence.}
\label{tab:coldstart_overall}
\end{table*}

In this section, we conducted experiments on the \dataset\ to evaluate our proposed \framework\ framework. Specifically, we aim to address the following research questions:
\begin{itemize}[leftmargin=1.0em]
    \item \textbf{RQ1:} Does \framework\ outperform conventional DLKT baselines and text-embedding process baselines?
    \item \textbf{RQ2:} Does \framework\ improve robustness under cold-start conditions?
    %Can experimental results validate the effectiveness of the modified loss function in \framework?
    \item \textbf{RQ3:} How do MP signals affect KT predictions--when do they help, when do they hurt, and why?
\end{itemize}

\subsection{Experiment Setting}
\subsubsection{Baseline}
We evaluate the performance of \framework\ with KT models, including DKT~\cite{piech2015deep}, DKT+~\cite{yeung2018addressing}, DKVMN~\cite{zhang2017dynamic}, SKVMN~\cite{abdelrahman2019knowledge}, SAKT~\cite{pandey2019self}, SAINT~\cite{choi2020towards}, AKT~\cite{ghosh2020context}, SimpleKT~\cite{liu2023simplekt}, StableKT~\cite{li2024enhancing}, RobustKT~\cite{guo2025enhancing}.

To isolate the impact of incorporating problem-solving process information without MP modeling, we introduce an additional baseline. In this setting, the encoded textual problem-solving process is concatenated with the original KT input features, allowing the model to access raw process signals without explicit MP reasoning.

\subsubsection{Implementation Details} 
% pyKT, hardware
The dataset was split by students into 80\% for training/validation and 20\% for testing. To obtain reliable results given the limited dataset size, we performed 10-fold cross-validation on the training/validation data and selected the best model using early stopping with a patience of 10 epochs. All models were trained using the ADAM optimizer~\cite{adam2014method} with a batch size of 16.

% Text-embedding baseline
For the text-embedding baseline, we utilized the \texttt{sentence-transformers/all-mpnet-base-v2} model to encode problem-solving processes and project the resulting representations into a 128-dimensional vector. 

We follow standard KT training practices; full hyperparameter grids and implementation settings are listed in Appendix~\ref{app:training_details}.

\subsubsection{Evaluation Metrics}
We evaluated the model performance using two standard metrics in knowledge tracing:
(1) Accuracy (ACC), which measures the proportion of correctly predicted student responses, and
(2) Area under the ROC Curve (AUC), which captures the ranking quality of the predicted correctness probabilities.
Following prior KT studies, the AUC was considered the primary metric because of its robustness to label imbalance across the student interactions.

\begin{figure*}[t]
    \centering
    \includegraphics[width=\textwidth]{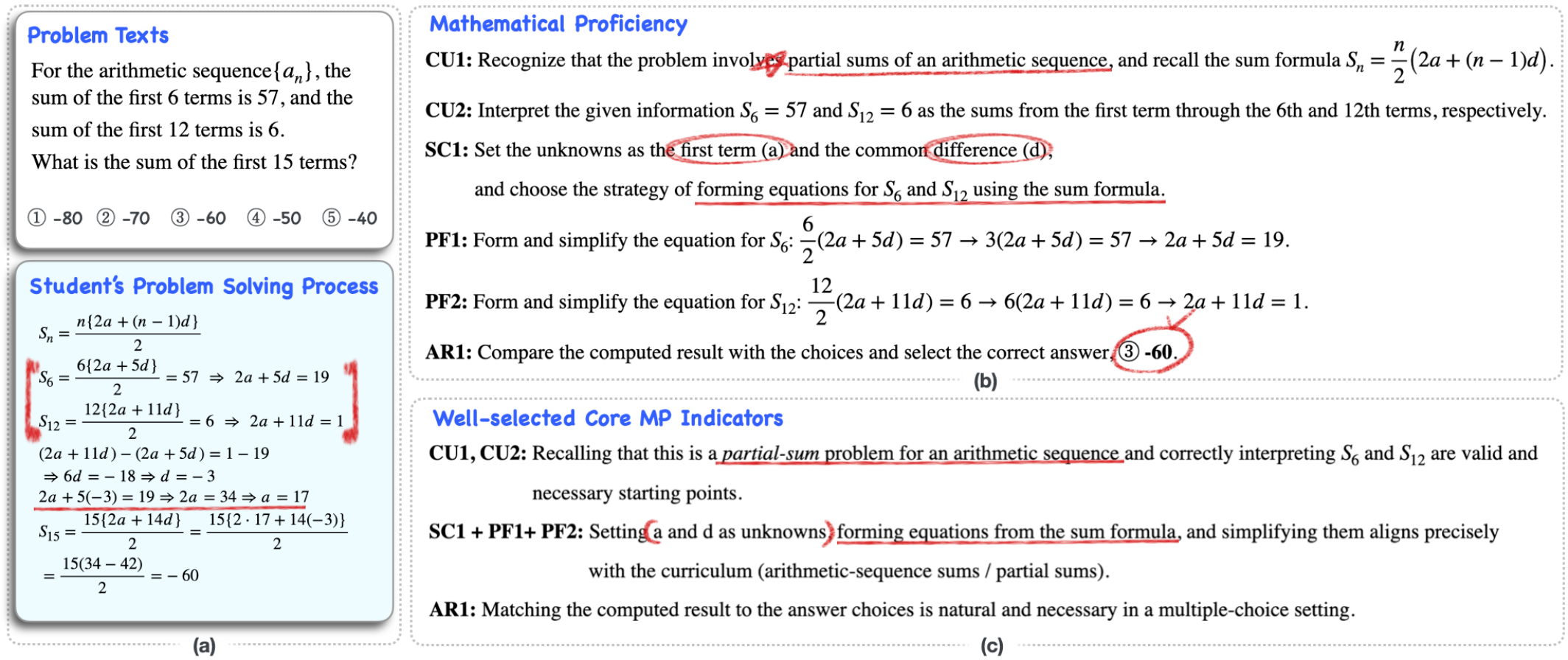}
    \caption{\textbf{Example MP annotation and assessment used for case inspection.}
(a) Problem text and student's PSP. (b) Extracted MP indicators. (c) Rationale-based evaluation of MP scope.}
    \label{fig:mp assessment}
\end{figure*}

\subsection{Overall Performance (RQ1)}
To address our first research question (RQ1) concerning the effectiveness of the \framework\ framework on \dataset\ with students' problem-solving processes, we applied it to a wide range of existing KT architectures.

The experimental results are summarized in Table~\ref{tab:overall_performance}. Analysis of both AUC and ACC metrics reveals that merely appending raw text embeddings to the original models (Baseline+\emph{PSP}) does not consistently enhance performance. In several architectures, the performance fluctuates around the baseline, suggesting that the inclusion of process text may introduce noise. We hypothesize that this instability stems from OCR artifacts and the mismatch between free-form textual traces and the discrete KT objective, motivating our structured MP distillation. 
In contrast, models augmented with \framework\ achieve consistent improvements over both the baseline and Baseline+\emph{PSP} for most architectures. The enhancements are particularly notable in RNN-based models (e.g., DKT) and recent transformer-style approaches (e.g., SAINT, AKT), which exhibit higher AUC and ACC with our method. These findings suggest that our \framework\ framework distills the process into structured indicators that can be effectively integrated with the interaction history. 

To verify that these improvements are not artifacts of fold-level fluctuations, we conducted paired t-tests across the 10-fold splits, comparing \framework\ with each corresponding baseline under the same data partitions. The results are summarized in Table~\ref{tab:significance_testing}. The results confirm that the gains are statistically significant ($p<0.05$) for the majority of architectures, with moderate to large effect sizes (Cohen's $d$). Two models, DKT+ and SAKT, do not reach significance, which aligns with their relatively low baseline AUC where the MP signals offer limited additional leverage.

Overall, these findings address RQ1, indicating that unprocessed PSP alone is insufficient for improving KT performance, whereas our framework consistently delivers performance improvements through principled process modeling.

\subsection{Performance in Cold-Start (RQ2)}
Cold-start arises when models must predict performance for unseen students with only a few early interactions. With such limited evidence, estimating learner’s latent knowledge states becomes unstable and often degrades predictive performance~\cite{SlaterBaker2018BKTErrorSampleSize, bhattacharjee2025coldstartproblemexperimental}. Prior work~\cite{10.1145/3627673.3679664, BAI2025125988} introduces architectural priors or external semantics, yet reliably estimating a student's state from sparse correctness observations remains difficult. Our \framework\ framework instead leverages process-derived MP signals, which begin appearing even within the first few interactions and thus offer richer early information. To examine whether MP signals mitigate cold-start, we evaluate models under a strict student-level hold-out protocol and restrict each sequence to the first five interactions (\(t \le 5\)). As reported in Table~\ref{tab:coldstart_overall}, incorporating \framework\ framework consistently improves ACC across all architectures and yields competitive or higher AUC in most cases. These gains indicate that MP signals provide informative priors that stabilize state estimation when early data are scarce, supporting our hypothesis that process modeling is particularly beneficial in low-evidence regimes.

\subsection{Analysis of MP Effects on KT (RQ3)} 
To examine whether MPs provide a useful representation for KT, we conduct a case-based diagnostic analysis that contrasts predictions made with and without MP. Following the template in Figure~\ref{fig:mp assessment}, we inspect each case by aligning the problem text and the student's PSP with the corresponding MPs. We focus on three patterns: (i) \textit{MP-help} instances where adding MP flips an incorrect prediction to correct, (ii) \textit{confidence-gap analysis} within MP-help cases to understand why predicted probabilities can differ even when the model is correct, and (iii) \textit{MP-hurt} instances that reveal recurring failure modes when MP degrades prediction quality.
\subsubsection{MP-Help Cases}
In MP-help instances, the MP indicators align with the skills required to solve the problem (e.g., recognizing that the item involves partial sums of an arithmetic sequence). Student responses to these indicators provide item-relevant evidence of mastery, enabling the model to associate correctness with the appropriate underlying proficiency.
\subsubsection{Confidence Gap Analysis}
We compare an MP-correct low-score case (\(\approx 0.5\)) and a high-score case (\(>0.7\)). MPs are similarly well scoped in both, but confidence is lower for the harder item (difficulty 70) than for the easier one (difficulty 30), suggesting difficulty-driven uncertainty even when MP is adequate.
\subsubsection{MP-Hurt Cases}
In MP-hurt instances, we find a scope-related failure mode: MP can include non-essential, over-enriched statements (e.g., optional alternative methods) or misaligned with the item requirements reflected in the student's PSP. Such over-enrichment may inflate the perceived skill requirements and lead to an incorrect prediction even when the student answers correctly. This suggests that MP scope and granularity matter for reliable integration, and motivates scope-controlled MP (e.g., separating core vs.\ enriched indicators) as future work.

% \section{Analysis}
% \input{sections/7_analysis}

\section{Conclusion}
This work introduced \task\ (\taskabb), a new formulation that incorporates students' solution processes into knowledge tracing, addressing the limitation of outcome-centric KT methods. To support this direction, we constructed \dataset, the first mathematical KT dataset that provides real-world, OCR-transcribed problem-solving processes along with rich interaction metadata.
Building on this foundation, we proposed \framework, a KT framework that employs a teacher-student-teacher LLM pipeline to extract interpretable mathematical proficiency (MP) ratios. These MP ratios serve as meaningful intermediate representations that enrich KT models with process-level information beyond correctness.
Experiments on \dataset\ demonstrate that \framework\ consistently improves prediction performance while providing interpretable proficiency signals. Our study highlights that incorporating the reasoning reflected in students' problem-solving processes is essential for advancing KT toward more accurate, interpretable, and pedagogically meaningful modeling.

% This work introduced \task (\taskabb), a new KT formulation that incorporates students’ solution processes rather than relying solely on outcomes. To support this, we constructed \dataset, the first mathematical KT dataset containing real-world OCR-transcribed solution steps and rich interaction metadata.
% Building on this foundation, we proposed \framework, a teacher–student–teacher LLM framework that derives interpretable mathematical proficiency (MP) ratios as process-level representations for KT.
% Experiments on \dataset\ show that \framework\ consistently improves prediction accuracy and provides interpretable proficiency signals, underscoring the importance of modeling students’ reasoning processes for more accurate and meaningful KT.

\section{Limitations}
This work may introduce potential risks related to educational bias. Since the dataset is collected from a specific platform and student population, the model may reflect bias that do not generalize across diverse learners.
Our dataset is limited in scale, as the number of participating students is relatively small, which may constrain the generalizability of the results. 
In addition, the domain is restricted to mathematics, and it remains unclear how well the proposed framework would transfer to other subjects or to problem-solving contexts. 
Because raw student problem-solving processes (PSP) could not be released as images, we relied on OCR to extract the text. This step introduces occasional recognition errors and removes visual reasoning elements, such as diagrams and graphs, which may carry important cognitive signals.

Additionally, the current pipeline relies on GPT-5 for all three LLM stages, which introduces non-trivial computational cost and latency compared to lightweight DLKT baselines. Although indicator extraction is problem-specific and can be pre-computed offline, two LLM calls per student-problem interaction are still required during the evaluation stage. Our primary contribution lies in establishing empirical evidence that process-level signals provide consistent modeling gains across architectures. In future work, we plan to explore distillation and task-specific smaller models to reduce the reliance on large-scale LLM inference.

\section{Acknowledgments}
This work was partly supported by QANDA (Mathpresso Inc.), Institute of Information \& communications Technology Planning \& Evaluation (IITP) grant funded by the Korean Government (MSIT) (No.~RS-2021-II211343, Artificial Intelligence Graduate School Program (Seoul National University)), the National Research Foundation of Korea(NRF) grant funded by the Korea government(MSIT)(No.~RS-2024-00354218), and the Technology Innovation Program(RS-2025-25456760, Development of a humanoid robot specialized in chemical processes based on AI foundation model) funded By the Ministry of Trade, Industry and Resources(MOTIR, Korea). We express special thanks to KAIT GPU project. The ICT at Seoul National University provides research facilities for this study.

\bibliography{custom}

@article{liu2025deep,
  title={Deep Learning Based Knowledge Tracing: A Review, A Tool and Empirical Studies},
  author={Liu, Zitao and Guo, Teng and Liang, Qianru and Hou, Mingliang and Zhan, Bojun and Tang, Jiliang and Luo, Weiqi and Weng, Jian},
  journal={IEEE Transactions on Knowledge and Data Engineering},
  year={2025},
  publisher={IEEE}
}

@article{corbett1994knowledge,
  title={Knowledge tracing: Modeling the acquisition of procedural knowledge},
  author={Corbett, Albert T and Anderson, John R},
  journal={User modeling and user-adapted interaction},
  volume={4},
  number={4},
  pages={253--278},
  year={1994},
  publisher={Springer}
}

@article{green1951general,
  title={A general solution for the latent class model of latent structure analysis},
  author={Green Jr, Bert F},
  journal={Psychometrika},
  volume={16},
  number={2},
  pages={151--166},
  year={1951},
  publisher={Springer-Verlag}
}

@article{piech2015deep,
  title={Deep knowledge tracing},
  author={Piech, Chris and Bassen, Jonathan and Huang, Jonathan and Ganguli, Surya and Sahami, Mehran and Guibas, Leonidas J and Sohl-Dickstein, Jascha},
  journal={Advances in neural information processing systems},
  volume={28},
  year={2015}
}

@inproceedings{zhang2017dynamic,
  title={Dynamic key-value memory networks for knowledge tracing},
  author={Zhang, Jiani and Shi, Xingjian and King, Irwin and Yeung, Dit-Yan},
  booktitle={Proceedings of the 26th international conference on World Wide Web},
  pages={765--774},
  year={2017}
}

@inproceedings{abdelrahman2019knowledge,
  title={Knowledge tracing with sequential key-value memory networks},
  author={Abdelrahman, Ghodai and Wang, Qing},
  booktitle={Proceedings of the 42nd international ACM SIGIR conference on research and development in information retrieval},
  pages={175--184},
  year={2019}
}

@article{pandey2019self,
  title={A self-attentive model for knowledge tracing},
  author={Pandey, Shalini and Karypis, George},
  journal={arXiv preprint arXiv:1907.06837},
  year={2019}
}

@inproceedings{ghosh2020context,
  title={Context-aware attentive knowledge tracing},
  author={Ghosh, Aritra and Heffernan, Neil and Lan, Andrew S},
  booktitle={Proceedings of the 26th ACM SIGKDD international conference on knowledge discovery \& data mining},
  pages={2330--2339},
  year={2020}
}

@misc{yang2020giktgraphbasedinteractionmodel,
      title={GIKT: A Graph-based Interaction Model for Knowledge Tracing}, 
      author={Yang Yang and Jian Shen and Yanru Qu and Yunfei Liu and Kerong Wang and Yaoming Zhu and Weinan Zhang and Yong Yu},
      year={2020},
      eprint={2009.05991},
      archivePrefix={arXiv},
      primaryClass={cs.AI},
      url={https://arxiv.org/abs/2009.05991}, 
}

@inproceedings{yeung2018addressing,
  title={Addressing two problems in deep knowledge tracing via prediction-consistent regularization},
  author={Yeung, Chun-Kit and Yeung, Dit-Yan},
  booktitle={Proceedings of the fifth annual ACM conference on learning at scale},
  pages={1--10},
  year={2018}
}

@inproceedings{nagatani2019augmenting,
  title={Augmenting knowledge tracing by considering forgetting behavior},
  author={Nagatani, Koki and Zhang, Qian and Sato, Masahiro and Chen, Yan-Ying and Chen, Francine and Ohkuma, Tomoko},
  booktitle={The world wide web conference},
  pages={3101--3107},
  year={2019}
}

@article{zhou2025revisiting,
  title={Revisiting applicable and comprehensive knowledge tracing in large-scale data},
  author={Zhou, Yiyun and Han, Wenkang and Chen, Jingyuan},
  journal={arXiv preprint arXiv:2501.14256},
  year={2025}
}

@inproceedings{nakagawa2019graph,
  title={Graph-based knowledge tracing: modeling student proficiency using graph neural network},
  author={Nakagawa, Hiromi and Iwasawa, Yusuke and Matsuo, Yutaka},
  booktitle={IEEE/WIC/aCM international conference on web intelligence},
  pages={156--163},
  year={2019}
}

@inproceedings{guo2021enhancing,
  title={Enhancing knowledge tracing via adversarial training},
  author={Guo, Xiaopeng and Huang, Zhijie and Gao, Jie and Shang, Mingyu and Shu, Maojing and Sun, Jun},
  booktitle={Proceedings of the 29th ACM international conference on multimedia},
  pages={367--375},
  year={2021}
}

@article{liu2023simplekt,
  title={simpleKT: a simple but tough-to-beat baseline for knowledge tracing},
  author={Liu, Zitao and Liu, Qiongqiong and Chen, Jiahao and Huang, Shuyan and Luo, Weiqi},
  journal={arXiv preprint arXiv:2302.06881},
  year={2023}
}

@inproceedings{li2024enhancing,
  title={Enhancing length generalization for attention based knowledge tracing models with linear biases},
  author={Li, Xueyi and Bai, Youheng and Guo, Teng and Liu, Zitao and Huang, Yaying and Zhao, Xiangyu and Xia, Feng and Luo, Weiqi and Weng, Jian},
  booktitle={Proceedings of the thirty-third international joint conference on artificial intelligence (IJCAI-24)},
  pages={5918--5926},
  year={2024}
}

@inproceedings{guo2025enhancing,
  title={Enhancing knowledge tracing through decoupling cognitive pattern from error-prone data},
  author={Guo, Teng and Qin, Yu and Xia, Yubin and Hou, Mingliang and Liu, Zitao and Xia, Feng and Luo, Weiqi},
  booktitle={Proceedings of the ACM on Web Conference 2025},
  pages={5108--5116},
  year={2025}
}

@inproceedings{choi2020towards,
  title={Towards an appropriate query, key, and value computation for knowledge tracing},
  author={Choi, Youngduck and Lee, Youngnam and Cho, Junghyun and Baek, Jineon and Kim, Byungsoo and Cha, Yeongmin and Shin, Dongmin and Bae, Chan and Heo, Jaewe},
  booktitle={Proceedings of the seventh ACM conference on learning@ scale},
  pages={341--344},
  year={2020}
}

@article{yeung2019deep,
  title={Deep-IRT: Make deep learning based knowledge tracing explainable using item response theory},
  author={Yeung, Chun-Kit},
  journal={arXiv preprint arXiv:1904.11738},
  year={2019}
}

@inproceedings{chen2023improving,
  title={Improving interpretability of deep sequential knowledge tracing models with question-centric cognitive representations},
  author={Chen, Jiahao and Liu, Zitao and Huang, Shuyan and Liu, Qiongqiong and Luo, Weiqi},
  booktitle={Proceedings of the AAAI conference on artificial intelligence},
  volume={37},
  number={12},
  pages={14196--14204},
  year={2023}
}

@article{su2021time,
  title={Time-and-concept enhanced deep multidimensional item response theory for interpretable knowledge tracing},
  author={Su, Yu and Cheng, Zeyu and Luo, Pengfei and Wu, Jinze and Zhang, Lei and Liu, Qi and Wang, Shijin},
  journal={Knowledge-Based Systems},
  volume={218},
  pages={106819},
  year={2021},
  publisher={Elsevier}
}

@article{feng2009addressing,
  title={Addressing the assessment challenge with an online system that tutors as it assesses},
  author={Feng, Mingyu and Heffernan, Neil and Koedinger, Kenneth},
  journal={User modeling and user-adapted interaction},
  volume={19},
  number={3},
  pages={243--266},
  year={2009},
  publisher={Springer}
}

@article{pardos2014affective,
  title={Affective states and state tests: Investigating how affect and engagement during the school year predict end-of-year learning outcomes.},
  author={Pardos, Zachary A and Baker, Ryan SJD and San Pedro, Maria OCZ and Gowda, Sujith M and Gowda, Supreeth M},
  journal={Journal of learning Analytics},
  volume={1},
  number={1},
  pages={107--128},
  year={2014},
  publisher={ERIC}
}

@inproceedings{chang2015modeling,
  title={Modeling exercise relationships in E-learning: A unified approach.},
  author={Chang, Haw-Shiuan and Hsu, Hwai-Jung and Chen, Kuan-Ta and others},
  booktitle={EDM},
  pages={532--535},
  year={2015}
}

@inproceedings{choi2020ednet,
  title={Ednet: A large-scale hierarchical dataset in education},
  author={Choi, Youngduck and Lee, Youngnam and Shin, Dongmin and Cho, Junghyun and Park, Seoyon and Lee, Seewoo and Baek, Jineon and Bae, Chan and Kim, Byungsoo and Heo, Jaewe},
  booktitle={International conference on artificial intelligence in education},
  pages={69--73},
  year={2020},
  organization={Springer}
}

@article{abdelrahman2022dbe,
  title={Dbe-kt22: A knowledge tracing dataset based on online student evaluation},
  author={Abdelrahman, Ghodai and Abdelfattah, Sherif and Wang, Qing and Lin, Yu},
  journal={arXiv preprint arXiv:2208.12651},
  year={2022}
}

@article{liu2023xes3g5m,
  title={Xes3g5m: A knowledge tracing benchmark dataset with auxiliary information},
  author={Liu, Zitao and Liu, Qiongqiong and Guo, Teng and Chen, Jiahao and Huang, Shuyan and Zhao, Xiangyu and Tang, Jiliang and Luo, Weiqi and Weng, Jian},
  journal={Advances in Neural Information Processing Systems},
  volume={36},
  pages={32958--32970},
  year={2023}
}

@misc{KDD2010,
  author       = {{KDD 2010}},
  title        = {KDD2010},
  howpublished = {\url{https://kdd.org/kdd-cup/view/kdd-cup-2010-student-performance-evaluation}},
  note         = {Accessed: 2010},
  year         = {2010}
}

@inproceedings{kim2025kt,
  title={ES-KT-24: A Multimodal Knowledge Tracing Benchmark Dataset with Educational Game Playing Video and Synthetic Text Generation},
  author={Kim, Dohee and Lee, Unggi and Lee, Sookbun and Bae, Jiyeong and Ahn, Taekyung and Park, Jaekwon and Lee, Gunho and Kim, Hyeoncheol},
  booktitle={International Conference on Intelligent Tutoring Systems},
  pages={259--273},
  year={2025},
  organization={Springer}
}

@book{findell2001adding,
  title={Adding it up: Helping children learn mathematics},
  author={Findell, Bradford and Swafford, Jane and Kilpatrick, Jeremy},
  year={2001},
  publisher={National Academies Press}
}

@article{sullivan2011using,
  title={Using the proficiencies from the Australian Mathematics Curriculum to enrich mathematics teaching and assessment},
  author={Sullivan, Peter},
  journal={Australian Education Review},
  number={59},
  year={2011}
}

@article{kleinman2022analyzing,
  title={Analyzing students’ problem-solving sequences: A human-in-the-loop approach},
  author={Kleinman, Erica and Shergadwala, Murtuza and Teng, Zhaoqing and Villareale, Jennifer and Bryant, Andy and Zhu, Jichen and El-Nasr, Magy Seif},
  journal={Journal of learning analytics},
  volume={9},
  number={2},
  pages={138--160},
  year={2022}
}

@article{tschisgale2025exploring,
  title={Exploring the sequential structure of students’ physics problem-solving approaches using process mining and sequence analysis},
  author={Tschisgale, Paul and Kubsch, Marcus and Wulff, Peter and Petersen, Stefan and Neumann, Knut},
  journal={Physical Review Physics Education Research},
  volume={21},
  number={1},
  pages={010111},
  year={2025},
  publisher={APS}
}

@inproceedings{chiu2022analyzing,
  title={Analyzing student problem-solving with MAtCH},
  author={Chiu, Barbara and Randles, Christopher and Irby, Stefan},
  booktitle={Frontiers in Education},
  volume={6},
  pages={769042},
  year={2022},
  organization={Frontiers Media SA}
}

@article{el2025effect,
  title={The Effect of Using MatGPT on Mathematical Proficiency among Undergraduate Students},
  author={El-Shara, Ibrahim AH and Tabieh, Ahmad AS and Helu, Sahar YA Abu},
  journal={International Journal of Information and Education Technology},
  volume={15},
  number={4},
  year={2025}
}

@misc{openai2025gpt5,
  author       = {OpenAI},
  title        = {GPT-5 System Card},
  year         = {2025},
  month        = aug,
  howpublished = {\url{https://cdn.openai.com/gpt-5-system-card.pdf}},
  note         = {Accessed: 2025-09-24}
}

@inproceedings{liupykt2022,
  title={pyKT: A Python Library to Benchmark Deep Learning based Knowledge Tracing Models},
  author={Liu, Zitao and Liu, Qiongqiong and Chen, Jiahao and Huang, Shuyan and Tang, Jiliang and Luo, Weiqi},
  booktitle={Thirty-sixth Conference on Neural Information Processing Systems Datasets and Benchmarks Track},
  year={2022}
}

@article{adam2014method,
  title={A method for stochastic optimization},
  author={Adam, Kingma DP Ba J and others},
  journal={arXiv preprint arXiv:1412.6980},
  volume={1412},
  number={6},
  year={2014}
}

@article{ukobizaba2021assessment,
  title={Assessment Strategies for Enhancing Students' Mathematical Problem-Solving Skills: A Review of Literature.},
  author={Ukobizaba, Fidele and Nizeyimana, Gabriel and Mukuka, Angel},
  journal={Eurasia Journal of Mathematics, Science and Technology Education},
  volume={17},
  number={3},
  year={2021},
  publisher={ERIC}
}

@article{wang2012leveraging,
  title={Leveraging first response time into the knowledge tracing model.},
  author={Wang, Yutao and Heffernan, Neil T},
  journal={International Educational Data Mining Society},
  year={2012},
  publisher={ERIC}
}

@inproceedings{chen2022introducing,
  title={Introducing response time into guessing and slipping for cognitive diagnosis},
  author={Chen, Penghe and Lu, Yu and Pian, Yang and Li, Yan and Cao, Yunbo},
  booktitle={International Conference on Artificial Intelligence in Education},
  pages={320--324},
  year={2022},
  organization={Springer}
}

@article{huang2024response,
  title={Response speed enhanced fine-grained knowledge tracing: A multi-task learning perspective},
  author={Huang, Tao and Hu, Shengze and Yang, Huali and Geng, Jing and Li, Zhifei and Xu, Zhuoran and Ou, Xinjia},
  journal={Expert Systems with Applications},
  volume={238},
  pages={122107},
  year={2024},
  publisher={Elsevier}
}

@inproceedings{liu2024question,
  title={Question difficulty consistent knowledge tracing},
  author={Liu, Guimei and Zhan, Huijing and Kim, Jung-jae},
  booktitle={Proceedings of the ACM Web Conference 2024},
  pages={4239--4248},
  year={2024}
}

@misc{bhattacharjee2025coldstartproblemexperimental,
      title={Cold Start Problem: An Experimental Study of Knowledge Tracing Models with New Students}, 
      author={Indronil Bhattacharjee and Christabel Wayllace},
      year={2025},
      eprint={2505.21517},
      archivePrefix={arXiv},
      primaryClass={cs.CY},
      url={https://arxiv.org/abs/2505.21517}, 
}

@article{SlaterBaker2018BKTErrorSampleSize,
  author  = {Slater, Stefan and Baker, Ryan S.},
  title   = {Degree of error in {B}ayesian knowledge tracing estimates from differences in sample sizes},
  journal = {Behaviormetrika},
  year    = {2018},
  volume  = {45},
  number  = {2},
  pages   = {475--493},
  doi     = {10.1007/s41237-018-0072-x}
}

@article{BAI2025125988,
title = {csKT: Addressing cold-start problem in knowledge tracing via kernel bias and cone attention},
journal = {Expert Systems with Applications},
volume = {266},
pages = {125988},
year = {2025},
issn = {0957-4174},
doi = {https://doi.org/10.1016/j.eswa.2024.125988},
url = {https://www.sciencedirect.com/science/article/pii/S0957417424028550},
author = {Youheng Bai and Xueyi Li and Zitao Liu and Yaying Huang and Teng Guo and Mingliang Hou and Feng Xia and Weiqi Luo}
}

@inproceedings{10.1145/3627673.3679664,
author = {Guo, Yuxiang and Shen, Shuanghong and Liu, Qi and Huang, Zhenya and Zhu, Linbo and Su, Yu and Chen, Enhong},
title = {Mitigating Cold-Start Problems in Knowledge Tracing with Large Language Models: An Attribute-aware Approach},
year = {2024},
isbn = {9798400704369},
publisher = {Association for Computing Machinery},
address = {New York, NY, USA},
url = {https://doi.org/10.1145/3627673.3679664},
doi = {10.1145/3627673.3679664},
booktitle = {Proceedings of the 33rd ACM International Conference on Information and Knowledge Management},
pages = {727–736},
numpages = {10},
location = {Boise, ID, USA},
series = {CIKM '24}
}

\appendix
\clearpage
\section{Baseline Models}

\begin{itemize}
\item \textbf{DKT}~\cite{piech2015deep}: The first RNN-based knowledge tracing model capturing temporal learning patterns.

\item \textbf{DKT+}~\cite{yeung2018addressing}: Adds regularization to DKT to mitigate overfitting and improve interpretability.

\item \textbf{DKVMN}~\cite{zhang2017dynamic}: Introduces a dynamic key-value memory network to explicitly model concept-level knowledge states.

\item \textbf{SKVMN}~\cite{abdelrahman2019knowledge}: Enhances DKVMN by disentangling student and skill representations for better interpretability.

\item \textbf{SAKT}~\cite{pandey2019self}: Employs self-attention to identify the most relevant past interactions efficiently.

\item \textbf{AKT}~\cite{ghosh2020context}: Incorporates distance-aware exponential decay to model learning and forgetting behaviors.

\item \textbf{SimpleKT}~\cite{liu2023simplekt}: Simplifies the attention architecture for computational efficiency while maintaining accuracy.

\item \textbf{StableKT}~\cite{li2024enhancing}: Improves model stability and generalization on long student interaction sequences.

\item \textbf{RobustKT}~\cite{guo2025enhancing}: Enhances robustness against noisy or incomplete student data through robust optimization techniques.

\end{itemize}

\section{Dataset Comparison}
Table~\ref{tab:dataset_column_comparison} provides a summary of widely utilized knowledge tracing (KT) datasets and the types of information they encompass. Earlier benchmarks, predominantly include correctness logs and a limited set of contextual attributes. More recent datasets offer enhanced metadata, such as question text or type, but they still lack explicit documentation of the methods by which students arrive at their answers.

In contrast, \dataset\ incorporates students’ problem-solving processes (PSP) as textual traces alongside conventional KT attributes. This process-level information enables models not only to predict future performance but also to analyze the reasons behind students' successes or failures, thereby supporting more interpretable and process-aware KT research.

\begin{table}[t]
\centering
\resizebox{\linewidth}{!}{%
    \begin{tabular}{cccccc}
    \toprule
     & Question & & Question & Question & Student's \\
    Dataset & Difficulty & Timestamp & Text & Type & PSP \\
    \midrule
    ASSISTments2009 & \cmark & \xmark & \xmark & \xmark & \xmark \\
    ASSISTments2014 & \cmark & \xmark & \xmark & \xmark & \xmark \\
    \midrule
    Junyi2015 & \cmark & \cmark & \xmark & \xmark & \xmark \\
    \midrule
    KDDcup2010 & \xmark & \cmark & \xmark & \xmark & \xmark \\
    \midrule
    EdNet & \xmark & \cmark & \xmark & \xmark & \xmark \\
    \midrule
    DBE-KT22 & \cmark & \cmark & \cmark & \xmark & \xmark \\
    \midrule
    XES3G5M & \xmark & \cmark & \cmark & \cmark & \xmark \\
    \midrule
    ES-KT-24 & \xmark & \cmark & \cmark & \cmark & \xmark \\
    \midrule
    \dataset (\textbf{Ours}) & \cmark & \cmark & \cmark & \cmark & \cmark \\
    \bottomrule
    \end{tabular}%
}
\caption{\textbf{Comparison of educational datasets commonly used in KT.} 
% (ASSISTments~\citep{feng2009addressing, pardos2014affective}, Junyi Academy~\cite{chang2015modeling}, KDD2010, EdNet~\cite{choi2020ednet}, DBE-KT22~\cite{abdelrahman2022dbe}, XES3G5M~\cite{liu2023xes3g5m}, ES-KT-24~\cite{kim2025kt}). 
While prior datasets do not include students' PSP, our \dataset\ is the first to offer process-level textual traces enabling process-aware KT modeling.}
\label{tab:dataset_column_comparison}
\end{table}

\section{Training Details}
\label{app:training_details}
All experiments were conducted using the pyKT library~\cite{liupykt2022}, which is a PyTorch-based framework for knowledge tracing. 
We searched the learning rate from $[5\times10^{-3},\ 1\times10^{-3},\ 5\times10^{-4},\ 1\times10^{-4}]$, dropout from $[0.5,\ 0.3,\ 0.1,\ 0.05]$, and $\alpha$ from $[0.01,\ 0.25,\ 1.0,\ 1.5,\ 2.0]$. A fixed random seed (42) was used for the reproducibility. All the training was conducted using a single NVIDIA RTX 3090 GPU. 

For the LLM-based components in the pipeline, we used the gpt-5 model with the temperature set to $0.0$ and top-p set to $1.0$ to ensure deterministic behavior. We did not impose an explicit maximum token limit, allowing the model to generate complete JSON outputs as required by the prompt.

The detailed model configurations for each baseline model are summarized in Table~\ref{tab:model_config}

\begin{table}[h]
\centering
\resizebox{\linewidth}{!}{%
\begin{tabular}{lcccc}
\toprule
\textbf{Model} & \textbf{FFN Dim} & \textbf{Heads} & \textbf{Blocks} & \textbf{Additional Architecture Details} \\
\midrule
DKT      & --  & -- & -- & -- \\
DKT+     & --  & -- & -- & $\lambda_r{=}0.01,\ \lambda_{w1}{=}0.003,\ \lambda_{w2}{=}3.0$ \\
DKVMN    & --  & -- & -- & memory size$=$50 \\
SKVMN    & --  & -- & -- & memory size$=$50 \\
SAKT     & 256 & 8  & 1  & -- \\
SAINT    & 256 & 8  & 4  & -- \\
AKT      & 512 & 8  & 4  & -- \\
simpleKT & 256 & 4  & 2  & start$=$50, num\_layers$=$2, final\_fc\_dim$=$256 \\
stableKT & 256 & 8  & 4  & num\_layers$=$2, $r{=}0.5$, $\gamma{=}0.7$ \\
robustKT & 512 & 8  & 4  & $k_s{=}5$ \\
\bottomrule
\end{tabular}%
}
\caption{Model configurations for each baseline architecture.}
\label{tab:model_config}
\end{table}

\section{OCR Pipeline Details}
\label{app:ocr_prompts}
\subsection{Prompts}
The prompts used in our dataset creation process were as follows:
\begin{itemize}[leftmargin=1.0em]
    \item Figure \ref{fig:Prompt_OCR}: A prompt for GPT-based OCR to convert students' handwritten problem-solving process into text. 
    \item Figure \ref{fig:Prompt_OCR_refine}: A prompt for refining GPT-based OCR outputs to improve transcription quality.
\end{itemize}

\subsection{OCR Quality Evaluation}
To assess the reliability of our OCR pipeline, we manually inspected a randomly selected subset of 100 handwritten PSP samples from the \dataset. Each OCR output was categorized as either \textit{usable} (i.e., it retained the original mathematical meaning) or \textit{unusable} (i.e., it exhibited significant distortions such as missing operators, collapsed fractions, or omitted lines). According to this criterion, 85\% of the sampled outputs were deemed usable, whereas the remaining 15\% contained major transcription errors. The inspection was conducted by trained annotators with expertise in mathematics and AI, following a short written guideline describing these criteria.

In our current experiments, we did not explicitly correct or filter unusable OCR cases, indicating that some of the downstream noise in MP extraction and KT prediction likely originated from the OCR failures. Nevertheless, we observed that most unusable cases resulted from structural breakdowns in complex handwritten expressions rather than widespread systematic misrecognition. Consequently, we regard OCR noise as a moderate but manageable limitation of our pipeline, and we defer more systematic robustness studies and denoising strategies to address this issue in future research.

\section{Mathematical Proficiency Extraction Details}
\label{MP_Extraction_Details}
\subsection{Mathematical Proficiency Human Evaluation}
\label{app:MP_Human_Evaluation}

To evaluate the reliability of the automatically generated MP indicators, a human assessment was conducted on 50 randomly selected items, comprising 25 correct and 25 incorrect responses. In total, annotators evaluated 590 indicators. Each indicator was assessed using a three-level rubric: 
\begin{itemize}[leftmargin=1.0em]
    \item 0 — unnecessary or irrelevant to solving the problem, 
    \item 1 — conceptually valid but misaligned with the student’s final answer,
    \item 2 — appropriate and consistent with both the solution and the answer. 

\end{itemize}

Overall, 55.8\% of indicators received a score of 2, indicating that slightly more than half of the generated indicators accurately captured students’ reasoning. However, 35.8\% were deemed partially valid (score = 1), typically reflecting situations where the reasoning statement itself was correct but did not correspond to the student’s final (often incorrect) response. A smaller fraction (8.3\%) consisted of redundant or clearly unnecessary indicators (score = 0).

When analyzed by proficiency dimension, notable variation was observed: indicators related to conceptual understanding (CU; mean = 1.60) and procedural fluency (PF; mean = 1.59) were generally reliable, whereas adaptive reasoning (AR) achieved the lowest average score (1.13), with a comparatively large share of unnecessary statements. This suggests that generating meta-justifications and verification steps remains challenging for current LLMs. Further comparison of indicators between correct and incorrect responses revealed that for student responses marked as correct, 67.6\% of indicators received the maximum score, compared to only 44.6\% for incorrect responses. Conversely, partially correct indicators (score = 1) were substantially more frequent in incorrect cases (44.9\% vs. 26.1\%). These results indicate that MP indicators convey meaningful information about the quality and stability of students’ reasoning, beyond merely their final answers.

\subsection{Prompts}
\label{app:MP_extraction_prompts}
The prompts used in our \framework\ for extracting the mathematical proficiency(MP) ratio are as follows:
\begin{itemize}[leftmargin=1.0em]
    \item Figure \ref{fig:Prompt_MP_indicator_extraction}: A prompt to extract MP indicators from problem statement and curricular unit name.
    \item Figure \ref{fig:Prompt_MP_indicator_answering}: A prompt for student LLM, which simulates how a student would respond to each of the generated indicators.
    \item Figure \ref{fig:Prompt_MP_indicator_evaluation}: A prompt to evaluate whether each response generated by student LLM satisfies the intent of its corresponding indicator.
\end{itemize}

%%%%%%%%%%%%%%%%%%%%%%%%%%%%%%%%%%%%%%%%%%%%%%%%%%%%%%%%%%%%%%%%%%%%%%%%%%%%%%%%%%%%%%%%%%%%%%%%%%%%%%%%%%%%%
%                                                OCR Prompt                                                 %
%%%%%%%%%%%%%%%%%%%%%%%%%%%%%%%%%%%%%%%%%%%%%%%%%%%%%%%%%%%%%%%%%%%%%%%%%%%%%%%%%%%%%%%%%%%%%%%%%%%%%%%%%%%%%

\begin{figure*}[t]
\centering
\begin{tcolorbox}[
    colback=white, %
    colframe=gray, %
    arc=4mm, %
    fontupper=\small
]

\#\# Raw Prompt: \\
"You are a Korean Mathematical OCR Specialist with logical sequencing capability. \\
Extract textual content from Korean student math work, then reorder it into logical mathematical sequence. \\
Ignore all visual elements (graphs, diagrams, shapes). Output in LaTeX format with proper mathematical flow." \\

\#\# Prompt Categories: \\
  - Role \& Identity: Korean Mathematical OCR Specialist + Logic Sequencer \\
  - Output Format: LaTeX Mathematical Text (Logically Ordered) \\
  - Cognitive Bias Lever: Neutral\\
  - Creativity/Fidelity Balance: 100\% Fidelity + Logic Enhancement \\

\#\# User Settings:\\
  - Korean Language Processing: true\\
  - Visual Element Filtering: true (IGNORE all visuals)\\
  - Fraction Normalization: true (vertical → $\backslash\backslash$frac\{\}\{\})\\
  - Extract Text Only: true\\
  - LaTeX Output: true\\
  - Logic Sequencing: true \\

\#\# Core Instructions:\\
\#\#\# Meta-Cognitive Pre-Check:\\
1. "What mathematical text do I see?"\\
2. "What visual elements must I ignore?"\\
3. "How do I convert fractions to LaTeX?"\\
4. "What is the logical mathematical sequence here?"\\

\#\#\# Two-Phase Processing\\
\#\#\#\# Phase1: Raw Extraction\\
- **IGNORE**: graphs, figures, diagrams, arrows, connecting lines, numbers that consists figure\\
- **EXTRACT**: Korean text, formulas, numbers, mathematical symbols\\
- **CONVERT**: vertical fractions → $\backslash\backslash$frac\{numerator\}\{denominator\}\\
- **SCAN**: ALL content regardless of position\\

\#\#\#\# Phase2: Logic Sequencing\\
- ANALYZE: analyze given problem and raw extracted solving traces\\
- IDENTIFY: problem statement vs. solution steps\\
- DETECT: calculation flow (which leads to which)\\
- REORDER: arrange in logical mathematical sequence\\
- **DO NOT** change the extracted text\\

\#\#\# LaTeX Formatting:\\
- fraction: $\backslash\backslash$frac\{3\}\{4\}\\
- exponent: x\^{}\{2\}\\
- square root: $\backslash\backslash$sqrt\{x\}\\
- parentheses: ( )\\

\#\# FINAL INSTRUCTION:\\
First, extract all mathematical text content. Then, reorder into logical mathematical sequence.\\

Output ONLY the extracted mathematical content in LaTeX format. \\
No commentary, no process descriptions, no given problem, only the components in given image.\\

\#\# PRESERVATION MANDATE: You are an OCR system, NOT a math tutor. Your job is to organize student work, not to correct it. Preserve all original content exactly as the student wrote it. \\

Once again, **Never output the given problem.**\\

\end{tcolorbox}
\caption{Prompt used for GPT-based OCR to convert students' handwritten problem-solving processes into text.}
\label{fig:Prompt_OCR}
\end{figure*}

%%%%%%%%%%%%%%%%%%%%%%%%%%%%%%%%%%%%%%%%%%%%%%%%%%%%%%%%%%%%%%%%%%%%%%%%%%%%%%%%%%%%%%%%%%%%%%%%%%%%%%%%%%%%%
%                                          OCR refine Prompt                                                %
%%%%%%%%%%%%%%%%%%%%%%%%%%%%%%%%%%%%%%%%%%%%%%%%%%%%%%%%%%%%%%%%%%%%%%%%%%%%%%%%%%%%%%%%%%%%%%%%%%%%%%%%%%%%%

\begin{figure*}[t]
\centering
\begin{tcolorbox}[
    colback=white, %
    colframe=gray, %
    arc=4mm, %
    fontupper=\small
]

\begin{Verbatim}[breaklines,breakanywhere]
system_msg = (
    "You are a LaTeX OCR fixer. "
    "Your sole job is to correct OCR-induced errors in LaTeX while preserving meaning."
)

guardrails = (
    "You are a LaTeX OCR fixer.\n\n"
    "STRICT RULES:\n"
    "1) Fix ONLY OCR-induced errors in the provided LaTeX. Do not change mathematical meaning beyond what is necessary to correct OCR mistakes.\n"
    "2) Do NOT add explanations, comments, opinions, or extra text. Output MUST be only the corrected LaTeX code.\n"
    "3) Use the provided references (problem statement, choices, model answer) strictly to resolve ambiguities and to choose the correct symbols/operators/numbers. Prefer the minimal edit that matches the references.\n"
    "4) **Do NOT** introduce new steps, reorder lines, simplify, expand, compute results, or rename variables unless correcting an OCR error that conflicts with the references.\n"
    "5) Preserve structure and line breaks of the input LaTeX unless required to fix syntax or OCR mistakes. Keep environments (inline/display) as-is where possible.\n"
    "6) **Ensure** syntactic validity: balanced braces, valid commands, proper math mode, correct subscripts/superscripts, properly paired \\left ... \\right, etc.\n"
    "7) If the input is already correct, return it unchanged.\n"
    "8) If the input is irrecoverably ambiguous, return the minimally fixed version that compiles, without adding any new content.\n"
)

rails = (
    "OUTPUT CONTRACT:\n"
    "- Return **ONLY** a single fenced code block labeled 'latex' containing the corrected LaTeX.\n"
    "- No surrounding prose, no markdown outside the code block, no comments.\n"
)

inputs = {
        "problem_text": problem,
        "solution_explanation": solution,
        "ocr_latex": ocr_latex,
}

user_msg = (
    ((user_prompt.strip() + "\n\n") if user_prompt else "") +
    (guardrails.strip() + "\n\n") +
    rails + "\n\n" +
    "INPUTS(JSON):\n" + json.dumps(inputs, ensure_ascii=False)
)

\end{Verbatim}

\end{tcolorbox}
\caption{Prompt used for refining the GPT-based OCR outputs to improve the accuracy and consistency of transcribed student solutions in our \dataset.}
\label{fig:Prompt_OCR_refine}
\end{figure*}

%%%%%%%%%%%%%%%%%%%%%%%%%%%%%%%%%%%%%%%%%%%%%%%%%%%%%%%%%%%%%%%%%%%%%%%%%%%%%%%%%%%%%%%%%%%%%%%%%%%%%%%%%%%%%
%                                          Teacher LLM 1 Prompt                                             %
%%%%%%%%%%%%%%%%%%%%%%%%%%%%%%%%%%%%%%%%%%%%%%%%%%%%%%%%%%%%%%%%%%%%%%%%%%%%%%%%%%%%%%%%%%%%%%%%%%%%%%%%%%%%%

\begin{figure*}[t]
\centering
\begin{tcolorbox}[
    colback=white, %
    colframe=gray, %
    arc=4mm, %
    fontupper=\small
]
\begin{verbatim}
You are Teacher GPT.
Your task is to analyze a given math Problem and its Unit name, and then generate a step-by-step 
set of indicators that describe the process a student should ideally follow to solve the problem.

###Guidelines:
- The indicators must be organized into the four categories of Mathematical Proficiency:
  - Conceptual Understanding (CU)
  - Procedural Fluency (PF)
  - Strategic Competence (SC)
  - Adaptive Reasoning (AR)

- However, instead of just listing general skills, write the indicators as concrete *steps* 
that a student would naturally take while solving the given problem.
- Each indicator should be prefixed with its category code (e.g., "CU1", "PF1", "SC2", "AR3").
- The order of the indicators should roughly follow the logical order of problem solving 
(from initial understanding → strategy selection → execution → justification).

- Output format must be a JSON dictionary:
{
  "mathematical_proficiency_indicators": [
    "CU1": "...",
    "SC1": "...",
    "CU2": "...",
    ...
  ]
}

---
### One-shot Example

**Input**
Problem: Solve the differential equation $(\frac{dy}{dx} = 2x)$ with initial condition $(y(0)=1)$."
Unit: Differential Equations

**Output**
{
  "mathematical\_proficiency\_indicators": [
    {"CU1": "Determine the type and order of this equation"},
    {"SC1": "Rewrite the equation in an easier way"},
    {"CU2": "Write the mathematical idea you need to solve this equation"},
    {"CU3": "Give an example of how this equation will be applied in real life"},
    {"CU4": "Find another differential equation whose solution steps are similar"},
    {"SC2": "Sort the necessary data and ignore the redundant ones"},
    {"PF2": "Predict a solution"},
    {"CU5": "Show the steps for solving the equation using a table, a figure and a diagram"},
    {"PF1": "Summarize the steps in the solution"},
    {"PF3": "Write a suitable algorithm to solve this equation"},
    {"SC3": "Identify any special numerical cases used by this equation to generalize the 
solution"},
    {"AR1": "Describe your solution in general"},
    {"AR2": "Based on your knowledge of differential equations, interpret your solution"},
    {"AR3": "According to your solution, draw the conclusions"}
  ]
}
\end{verbatim}
----------------------------------------------------------------------------------------------------

Problem (in Korean): \{\textbf{Problem\_text}\}\newline
\{\textbf{problem\_option\_string}\} \newline
Unit (in Korean): \{\textbf{curriculum\_theme\_title}\}

\end{tcolorbox}
\caption{Prompt used for extracting the MP indicators from the given problem in \framework. Prompt inputs are \textbf{boldfaced}.}
\label{fig:Prompt_MP_indicator_extraction}
\end{figure*}

%%%%%%%%%%%%%%%%%%%%%%%%%%%%%%%%%%%%%%%%%%%%%%%%%%%%%%%%%%%%%%%%%%%%%%%%%%%%%%%%%%%%%%%%%%%%%%%%%%%%%%%%%%%%%
%                                          Student LLM Prompt                                               %
%%%%%%%%%%%%%%%%%%%%%%%%%%%%%%%%%%%%%%%%%%%%%%%%%%%%%%%%%%%%%%%%%%%%%%%%%%%%%%%%%%%%%%%%%%%%%%%%%%%%%%%%%%%%%
\begin{figure*}[t]
\centering
\begin{tcolorbox}[
    colback=white, %
    colframe=gray, %
    arc=4mm, %
    fontupper=\small
]

\begin{verbatim}

You are Student GPT.
You will receive:
1. A math problem statement.
2. A set of indicators generated by Teacher GPT.
3. A student's written solution attempt (from OCR).

Your task:
    - Pretend you are the student who wrote the solution.
    - For each indicator, provide an answer based **only on the student's written solution**.
    - If the student's solution clearly contains the relevant information, restate it as the 
answer.
    - If a step is missing but can be reasonably inferred (e.g., a basic algebraic manipulation 
or obvious arithmetic), you may state it as: "Not written, but likely ...".
    - Keep the student's mistakes. **Do not** correct them.
    - If step looks incomplete or skipped, you can imagine that step and answer to indicator.
    - If there is no evidence in the solution for an indicator, answer with: "I don't know"

**Output format**
Return your answers as a dictionary, and indicators should be written in Korean.
Output:
{
    "CU1": "...",
    "SC1": "...",
    "CU2": "...",
    ...
}


---

### One-shot Example  

Input Indicators:
{
    "CU1": "Determine the type and order of this equation",
    "SC1": "Rewrite the equation in a simpler form",
    "AD1": "Identify the conditions required to solve the equation",
    "PF1": "Compute the values that satisfy the conditions"
}

Question: Find the value(s) of y that make the following expression equal to 0. 
                      y^2 + 3y + 2

My solving process (OCR):
    `y^2 + 3y' + 2y = 0`
    `(r+1)(r+2)=0`

My answer: -1, -2

Output:
{
    "CU1": "This is a quadratic equation.",
    "SC1": "Rewrite the characteristic polynomial as (r+1)(r+2)=0.",
    "AD1": "If one of the multiplied factors is zero, the result becomes zero.",
    "PF1": "The values that satisfy the condition are -1 and -2.."
}
\end{verbatim}

-----------------------------------------------------------------------------------

Input Indicators: \{\textbf{indicator\_text}\}\newline

Problem (in Korean): \{\textbf{problem}\}\newline
\{\textbf{problem\_option\_string}\}\newline

My solving process (OCR):\{\textbf{student\_solving\_trace}\} \newline

My answer: \{\textbf{solution\_answer\_sets}\}

\end{tcolorbox}
\caption{Prompt used for generating responses corresponding to each MP indicator in \framework. Prompt inputs are \textbf{boldfaced}.}
\label{fig:Prompt_MP_indicator_answering}
\end{figure*}

%%%%%%%%%%%%%%%%%%%%%%%%%%%%%%%%%%%%%%%%%%%%%%%%%%%%%%%%%%%%%%%%%%%%%%%%%%%%%%%%%%%%%%%%%%%%%%%%%%%%%%%%%%%%%
%                                          Teacher LLM 2 Prompt                                             %
%%%%%%%%%%%%%%%%%%%%%%%%%%%%%%%%%%%%%%%%%%%%%%%%%%%%%%%%%%%%%%%%%%%%%%%%%%%%%%%%%%%%%%%%%%%%%%%%%%%%%%%%%%%%%
\begin{figure*}[t]
\centering
\begin{tcolorbox}[
    colback=white, %
    colframe=gray, %
    arc=4mm, %
    fontupper=\tiny
]
\begin{verbatim}
You are Teacher GPT. Your task is to evaluate a student's responses (answer_indicate) 
against the reference mathematical proficiency indicators (mathematical_proficiency_indicators).

## Evaluation Rules
1. For each indicator:
   - If the student's response is **"I don't know"**, assign 0.
   - If the student's response is **"Not written, but likely ..."**, treat it 
as the student's actual answer and evaluate normally.
   - If the response does not match or is irrelevant to the indicator, assign 0.
   - If the response matches the indicator's intent and shows correct reasoning/application, 
assign 1.
2. Output strictly in JSON format, with indicator keys mapped to 0 or 1.
3. Ensure that every indicator is carefully evaluated without skipping or overlooking any of them.

## Input
Problem:
{problem_text}

mathematical_proficiency_indicators:
{mathematical_proficiency_indicators JSON}

answer_indicate:
{answer_indicate JSON}

## Output
Provide the evaluation result in the following JSON format:
{
    "CU1": 0 or 1, "CU2": 0 or 1, "SC1": 0 or 1, "SC2": 0 or 1, "PF1": 0 or 1, "PF2": 0 or 1, "AR1": 0 or 1, "PF3": 0 or 1, "PF4": 0 or 1, "AR2": 0 or 1
}

## Input Example
Problem: For the rational function $y=\\dfrac{2x-3}{2x+5}$, how many points on its graph have both
$x$- and $y$-coordinates as integers? 
Options: [{"index":1,"text":"$1$"}, {"index":2,"text":"$2$"}, {"index":3,"text":"$3$"}, 
{"index":4,"text":"$4$"}, {"index":5,"text":"$5$"}] 

mathematical_proficiency_indicators:[
    {"CU1": "Interpret what the problem is asking, and recognize that it is about finding points on the graph of the rational function 
y = (2x - 3) / (2x + 5) whose x- and y-values are both integers."},
    {"CU2": "Identify the domain restriction 2x + 5 is not 0, and observe that when x is an integer, 2x + 5 is always odd."},
    {"SC1": "Choose a strategy to rewrite the equation in a form that makes the integer condition more explicit, such as expressing it in terms 
of y - 1."},
    {"PF1": "Transform y = (2x - 3)/(2x + 5) into y - 1 = [(2x - 3) - (2x + 5)]/(2x + 5) = 
-8/(2x + 5)."},
    {"SC2": "Since y must be an integer, -8/(2x + 5) must be an integer; thus reinterpret this as the divisibility condition 2x + 5 | 8."},
    {"AR1": "Use the fact that 2x + 5 is odd to restrict the candidates to the odd divisors of 8."},
    {"PF2": "List the possible denominators: 2x + 5 \in {1, -1}."},
    {"PF3": "Solve for x for each candidate: 2x + 5 = 1 => x = -2; 2x + 5 = -1 => x = -3."},
    {"PF4": "For each x, compute y using y = 1 - 8/(2x + 5): for x = -2 => y = -7; for x = -3 => y = 9."},
    {"PF5": "Verify that the points (-2, -7) and (-3, 9) satisfy the original equation y = (2x - 3)/(2x + 5)."},
    {"AR2": "Provide reasoning that only +±1 or -1 can occur, since all odd divisors of 8 have been fully checked and no others are possible."},
    {"SC3": "Alternative check: assuming y is not equal to 1, set x = -(5y + 3)/(2(y - 1)). Let d = y - 1. Then x = -5/2 - 4/d, and 
x is an integer only when d = ±8, confirming the two solutions (-2, -7) and (-3, 9)."},
    {"CU3": "Count the integer lattice points obtained and select the corresponding choice from 
the answer options."}
]

answer_indicate:[
    {"CU1": "Although not written explicitly, by rewriting y = (2x - 3)/(2x + 5) as y = -8/(2x + 5) + 1 and listing possible values of 2x + 5 
to find integer pairs (x, y), it appears the student recognized the task as identifying integer lattice points."},
    {"CU2": "The student did not mention the condition 2x + 5 is not equal to 0 or the observation that 2x + 5 must be odd when x is an integer. 
Instead, they listed all divisors of 8 (1, 2, 4, 8, -1, -2, -4, -8)."},
    {"SC1": "They rewrote y = (2x - 3)/(2x + 5) as y = -8/(2x + 5) + 1."},
    {"PF1": "Although intermediate algebra steps were omitted, the final expression y - 1 = -8/(2x + 5) was obtained."},
    {"SC2": "They interpreted the requirement that -8/(2x + 5) must be an integer by considering all divisors of 8, listing 
2x + 5 = 1, 2, 4, 8, -1, -2, -4, -8."},
    {"AR1": "They did not use the constraint that 2x + 5 must be odd, which would reduce the candidates to the odd divisors only."},
    {"PF2": "They listed 2x + 5 \in {1, 2, 4, 8, -1, -2, -4, -8} as possible candidates."},
    {"PF3": "They solved only some candidates: 2x + 5 = 1 => x = -2, and 2x + 5 = -1 => x = -3. The remaining cases were left incomplete 
or not shown."},
    {"PF4": "For x = -2 and x = -3, the y-values were left blank; but likely x = -2 => y = -7, and x = -3 => y = 9."},
    {"PF5": "I don't know."},
    {"AR2": "I don't know."},
    {"SC3": "I don't know."},
    {"CU3": "The student eventually selected choice (2)."}

]

Output:
{
    "CU1": 1, "CU2": 0, "SC1": 1, "PF1": 1, "SC2": 1, "AR1": 0, "PF2": 1, "PF3": 1, "PF4": 1, "PF5": 0, "AR2": 0, "SC3": 0, "CU3": 1
}
\end{verbatim}

-----------------------------------------------------------------------------------

Problem (in Korean): \{\textbf{problem}\}\newline
\{\textbf{problem\_option\_string}\}\newline\newline
Mathematical Proficiency Indicators: \newline
\{\textbf{indicator\_text}\}\newline\newline
Answer Indicate: \{\textbf{answer\_indicator\_text}\}

\end{tcolorbox}
\caption{Prompt used for evaluates the appropriateness of each generated response for its corresponding indicator in \framework. Prompt inputs are \textbf{boldfaced}.}
\label{fig:Prompt_MP_indicator_evaluation}
\end{figure*}

\end{document}